\documentclass[sigconf]{acmart}

\usepackage{hyperref}
\usepackage{float}
\usepackage{dblfloatfix}
\usepackage{url}
\usepackage{amsmath}
\usepackage{bm}
\usepackage{graphicx}
\usepackage{algpseudocode}
\usepackage{algorithm}
\DeclareMathOperator*{\argmax}{arg\,max}

\newcommand{\mbf}[1]{\mathbf{#1}}
\newcommand{\bv}[1]{\mathbf{#1}}

\AtBeginDocument{%
  \providecommand\BibTeX{{%
    \normalfont B\kern-0.5em{\scshape i\kern-0.25em b}\kern-0.8em\TeX}}}

\copyrightyear{2020}
\acmYear{2020}
\setcopyright{acmlicensed}\acmConference[KDD '20]{Proceedings of the 26th ACM SIGKDD Conference on Knowledge Discovery and Data Mining}{August 23--27, 2020}{Virtual Event, CA, USA}
\acmBooktitle{Proceedings of the 26th ACM SIGKDD Conference on Knowledge Discovery and Data Mining (KDD '20), August 23--27, 2020, Virtual Event, CA, USA}
\acmPrice{15.00}
\acmDOI{10.1145/3394486.3403185}
\acmISBN{978-1-4503-7998-4/20/08}

\begin{document}
\fancyhead{}

\title{InfiniteWalk: Deep Network Embeddings as Laplacian Embeddings with a Nonlinearity}


\author{Sudhanshu Chanpuriya}
\affiliation{%
  \institution{University of Massachusetts Amherst}
  \city{Amherst}
  \state{Massachusetts}
  \postcode{01002}
}

\author{Cameron Musco}
\affiliation{%
  \institution{University of Massachusetts Amherst}
  \city{Amherst}
  \state{Massachusetts}
  \postcode{01002}
}

\renewcommand{\shortauthors}{Chanpuriya and Musco}

\begin{abstract}

  The skip-gram model for learning word embeddings \cite{mikolov2013distributed} has been widely popular, and DeepWalk~\cite{perozzi2014deepwalk}, among other methods, has extended the model to learning node representations from networks. Recent work of Qiu et al.~\cite{qiu2018network} provides a closed-form expression for the DeepWalk objective, obviating the need for sampling for small datasets and improving accuracy. In these methods, the ``window size" $T$ within which words or nodes are considered to co-occur is a key hyperparameter. We study the objective in the $T\to \infty$ limit, which allows us to simplify the expression from \cite{qiu2018network}. We prove that this limiting objective corresponds to factoring a simple transformation of the pseudoinverse of the graph Laplacian, linking DeepWalk to extensive prior work in spectral graph embeddings. 
  
  Further, we show that by applying a simple nonlinear entrywise transformation to this pseudoinverse, we recover a  good approximation of the finite-$T$ objective and embeddings that  
 %
 are competitive with those from DeepWalk and other skip-gram methods in multi-label classification. Surprisingly, we find that even simple binary thresholding of the Laplacian pseudoinverse is often competitive, suggesting that the core advancement of recent methods is a nonlinearity on top of the classical spectral embedding approach.
\end{abstract}

\ccsdesc[500]{Information systems~Data mining}
\ccsdesc[500]{Computing methodologies~Dimensionality reduction and manifold learning}

\keywords{spectral clustering, node embeddings, node classification, graph Laplacian, DeepWalk}


\maketitle

\section{Introduction}
Vertex embedding is the task of learning representations of graph vertices in a continuous vector space for use in downstream tasks, such as link prediction and vertex classification \cite{yan2006graph}.
The primary classical approach for this task is spectral embedding: vertices are represented by their corresponding values in the smallest eigenvectors of the graph Laplacian. Spectral embedding methods include the Shi-Malik normalized cuts algorithm~\cite{shi2000normalized}, Laplacian Eigenmaps~\cite{belkin2003laplacian}, and spectral partitioning methods for stochastic block models~\cite{mcsherry2001spectral}. They also include many spectral clustering methods, which apply spectral embeddings to general datasets by first transforming them into a graph based on data point similarity \cite{ng2002spectral}.

The spectral embedding approach has recently been exceeded in predictive performance on many tasks by methods inspired by Word2vec~\cite{mikolov2013distributed}, which performs the related word embedding task.
Word2vec forms representations for words based on the frequency with which they co-occur with other words, called context words, within a fixed distance $T$ in natural text. The DeepWalk~\cite{perozzi2014deepwalk}, LINE~\cite{tang2015line}, and node2vec~\cite{grover2016node2vec} algorithms, among others, adapt this idea to network data. In particular, DeepWalk takes several random walks on the network, treating the vertices as words, and treating the walks as sequences of words in text.
It has been shown in \cite{levy2014neural} that the representations learned by this approach are implicitly forming a low-dimensional factorization of a known matrix, which contains the pointwise mutual information (PMI) of co-occurences between nodes in the random walks.
Work by Qiu et al.~\cite{qiu2018network} gives  a closed form expression for this matrix and shows a connection to the normalized graph Laplacian. This motivates their NetMF algorithm, which performs a direct factorization, improving on the performance of DeepWalk on a number of tasks.

In this work, we consider DeepWalk in the limit as the window size $T$ goes to infinity.
We derive a simple expression for the PMI matrix in this limit:
%
\begin{align}\label{eq:simple}
\bv{M_\infty} = v_G \cdot \mbf{D}^{-1/2} \left( \mbf{\tilde{L}}^+ - \mbf{I} \right) \mbf{D}^{-1/2} + \mbf{J},
\end{align}
where $\mbf{D}$ is the degree matrix, $v_G$ is the trace of $\mbf{D}$ (twice the number of edges in $G$), $\mbf{\tilde{L}}$ is the normalized Laplacian (i.e. $\mbf{I} - \mbf{D}^{-1/2}\mbf{A} \mbf{D}^{-1/2}$), and $\mbf{J}$ is the all-ones matrix. $\mbf{\tilde{L}}^+$ is the pseudoinverse of $\mbf{\tilde{L}}$. One can  show that $\mbf{D}^{-1/2}\mbf{\tilde{L}}^+ \mbf{D}^{-1/2}$ is equal to the unnormalized Laplacian pseudoinverse $\mbf{L}^+$, the central object in classic spectral embeddings, up to a rank-2 component (see \eqref{eq:rank3} in Section \ref{sec:binary}). Thus, $\bv{M_\infty}$ is equal to $\mbf{L}^+$ plus at most a rank-3 component and a diagonal matrix.

Not surprisingly, embeddings formed directly using a low dimensional factorization of $\bv{M_\infty}$ itself perform poorly on downstream tasks compared to DeepWalk and other skip-gram methods. However, we show that after an element-wise application of a linear function followed by a logarithm, in particular, $x \to \log(1+x/T)$, $\bv{M_\infty}$ approximates the finite-$T$ PMI matrix. Embeddings formed by factorizing this transformed matrix are competitive with DeepWalk and nearly competitive with the NetMF method of Qiu et al \cite{qiu2018network}, when evaluated on multi-label node classification tasks. We call our method InfiniteWalk.

Note that the window hyperparameter $T$ only appears in the entrywise nonlinearity in InfiniteWalk and not in the formula for $\bv{M_\infty}$. 
This is perhaps surprising, as $T$ is a key hyperparameter in existing methods. Our results suggest that $T$'s importance lies largely in determining the shape of the nonlinearity applied. Since $\bv{M_\infty}$ is closely related to the Laplacian pseudoinverse, the key difference between DeepWalk and classic spectral embedding seems to be the application of this nonlinearity. 

In more detail, note that our results show that InfiniteWalk well approximates DeepWalk by forming a low-rank factorization of a nonlinear entrywise transformation of $\bv{M_\infty}$. Classic spectral embedding and clustering methods \cite{shi2000normalized,mcsherry2001spectral,belkin2003laplacian} embed nodes using the eigenvectors corresponding to the smallest eigenvalues of the Laplacian $\bv{L}$ (or a variant of this matrix), which are the largest eigenvalues of $\bv{L}^+$. Thus, these methods can be viewed as  embedding nodes using an optimal low-dimensional factorization of $\bv{L}^+$ (lying in the span of $\bv{L}^+$'s top eigenvectors), without applying an entrywise nonlinearity. 

Inspired by this observation, we simplify the idea of a nonlinear transformation of the Laplacian pseudoinverse even further: thresholding it to a binary matrix. In particular, we form the binary matrix
\[ [\mbf{L}^+ \geq c] ,\]
where $c$ is an element of $\mbf{L}^+$ itself of some hyperparameter quantile (e.g. the median or the $95$\textsuperscript{th} percentile element). Surprisingly, embeddings from factorizing this binary matrix are also competitive with DeepWalk and the method of Qiu et al. on many tasks.

Broadly, our results strengthen the theoretical connection between classical methods based on factorizing the graph Laplacian and more recent ``deep" methods for vertex embedding. They suggest that these methods are not as different conceptually as they may seem at first glance.

\section{Background and Related Work}

We begin by surveying existing work on skip-gram-based network embeddings and their connections to matrix factorization, which our work builds on.

\subsection{Skip-Gram} In word embedding models, words are typically treated as both words and contexts. A context is simply a word that appears within a window of length $T$ of another word. As formalized in~\cite{goldberg2014word2vec}, given a dataset of words $w \in \mathcal{W}$, contexts $c \in \mathcal{C}$ (typically $\mathcal{W} = \mathcal{C}$), and $(w, c)$ word-context co-occurrence pairs $(w, c) \in \mathcal{D}$, the ``skip-gram" model for training word and context embeddings $\mbf{v}_w$, $\mbf{v}_c$ \cite{mikolov2013distributed} has the following log-likelihood objective:
\[ \argmax_{\{\mbf{v}_c\}_{\mathcal{C}},\{\mbf{v}_w\}_{\mathcal{W}}} \sum_{(w,c) \in  \mathcal{D}} \log \Pr(c|w)
\qquad \Pr(c|w) = \frac{e^{\mbf{v}_c \cdot \mbf{v}_w}}{\sum_{c' \in \mathcal{C}} e^{\mbf{v}_{c'} \cdot \mbf{v}_w}} .\]
We can see that the objective encourages co-occuring pairs $(w,c) \in \mathcal{D}$ to have similar embeddings with large dot  product $\mbf{v}_c \cdot \mbf{v}_w$.
This exact objective is not used as repeatedly evaluating the partition function is too computationally expensive; \cite{mikolov2013distributed} proposes a substitute: the skip-gram with negative sampling (SGNS).
%
%
Here, an auxiliary `negative' dataset $\mathcal{D'}$ consisting of random $(w,c)$ pairs not appearing in $\mathcal{D}$ is used. Pairs in this negative set are encouraged to have dissimilar embeddings  with small $\mbf{v}_c \cdot \mbf{v}_w$.

\subsection{Implicit PMI Matrix} Levy and Goldberg~\cite{levy2014neural} prove that SGNS implicitly factors a matrix whose entries gave the pointwise mutual information (PMI) of occurrence of a word $w_i$ and occurrence of a context $c_j$. Given a dataset of these co-occurrences, an element of this matrix $\mbf{M}$ is given by
\begin{align*}
    \mbf{M}_{ij} &= \log \left( \frac{\Pr(w_i,c_j)}{\Pr(w_i)\Pr(c_j)} \right) - \log(b) \\
    &= \log \left( \frac{\#(w,c) \cdot |\mathcal{D}|}{\#(w) \cdot \#(c)} \right) - \log(b) .
\end{align*}
where $b = |\mathcal{D'}| / |\mathcal{D}|$ is the ratio of negative samples to positive samples. Their proof assumes that the dimensionality of the embedding is at least the cardinality of the word/context set; the analysis of Li et al.~\cite{li2015word} relaxes assumptions under which this equivalence holds. Several works, including \cite{arora2016latent} and \cite{gittens2017skip}, propose generative models for word-context co-occurrence to explain the effectiveness of PMI-based methods in linearizing semantic properties of words. More recently, the analysis of Allen et al. in \cite{allen2019vec} and \cite{allen2019analogies} has provided explanations of this phenomenon based on geometric properties of the PMI matrix, without the strong assumptions required by the generative models.
The extensive research and results on the skip-gram PMI matrix make it an intrinsically interesting object in representation learning.

\subsection{Networks} The DeepWalk method \cite{perozzi2014deepwalk} applies the SGNS model to networks, where the word and context sets are the nodes in the network, and the co-occuring pairs $\mathcal{D}$ are node pairs  that appear within a window of length $T$ hops in a set of length $L$ random walks run on the graph.
Qiu et al.~\cite{qiu2018network} derived the following expression for the PMI matrix in the context of random walks on undirected, connected, and non-bipartite graphs for DeepWalk: in the limit as the number of walks originating at each node $\gamma \to \infty$ and walk length $L\to \infty$, it approaches
\begin{align}\label{eq:netMF} \mbf{M_T} = \log\left( v_G \left( \tfrac{1}{T} \sum_{k=1}^T \mbf{P}^k \right) \mbf{D}^{-1} \right) - \log{b} , 
\end{align}
where $\log$ is an element-wise natural logarithm, $v_G$ (the ``volume" of the graph) is the sum of the degrees of all of the vertices, and $\mbf{P}=\mbf{D}^{-1} \mbf{A}$ is the random walk transition matrix.

Rather than sampling from random walks as in DeepWalk, the NetMF algorithm of Qiu et al. explicitly calculates and factors this matrix to produce embeddings, outperforming DeepWalk significantly on multi-label vertex classification tasks.
For low $T$, NetMF manually computes the exact sum, whereas for high $T$, it computes a low-rank approximation via SVD of the symmetrized transition matrix, $\mbf{\tilde{P}} = \mbf{D}^{-1/2}\mbf{A}\mbf{D}^{-1/2}$. While Qiu et al. analyze the effect of increasing $T$ on the spectrum of the resulting matrix, they do not pursue the $T \to \infty$ limit, stopping at $T=10$ as in the original DeepWalk paper. We show that this limiting matrix is both meaningful and simple to express.

\subsection{Other Approaches} Some other node embedding algorithms share significant similarities with DeepWalk. Qiu et al~\cite{qiu2018network} showed the LINE method to also be implicit matrix factorization, though its algorithm is based on edge sampling rather than sampling from random walks. In particular, its factorized matrix is a special case of the DeepWalk matrix with $T=1$. We include the performance of LINE in our empirical results.
node2vec~\cite{grover2016node2vec} is a generalization of DeepWalk which uses second-order random walks: the distribution of the following node in node2vec walks depends on the current and preceding nodes rather than only the current node as in DeepWalk. Hyperparameters allow the walk to approach BFS-like or DFS-like behavior as desired, which the authors assert extract qualitatively different information about node similarities.

Several architectures for applying convolutional neural networks to network data in an end-to-end fashion have been developed in the past few years, including the graph convolutional networks (GCNs) of \cite{kipf2016semi} and \cite{defferrard2016convolutional}, and some methods leverage these architectures to produce node embeddings: for example, Deep Graph Infomax~\cite{velivckovic2018deep} uses GCNs to maximize a mutual information objective involving patches around nodes. Recent work from Wu et al.~\cite{wu2019simplifying} shows that much of the complexity of GCNs comes from components inspired by other forms of deep learning that have limited utility for network data. In the same way, we seek to further the investigation of the core principles of ``deep" network embeddings apart from their inspiration in word embedding and neural networks. We note that, like DeepWalk, and the related methods, we focus on unsupervised embeddings, derived solely from a graph's structure, without training, e.g., on vertex labels.

\section{Methodology}

We now present our main contributions, which connect DeepWalk in the infinite window limit to classic spectral embedding with a nonlinearity. We discuss how this viewpoint clarifies the role of the window size parameter $T$ in DeepWalk and motivates a very simple embedding technique based on a binary thresholding of the graph Laplacian pseudoinverse.

\subsection{Derivation of Limiting PMI Matrix}
We start  by showing how to simplify the expression in \eqref{eq:netMF} for the DeepWalk PMI  given by \cite{qiu2018network} in the limit $T \rightarrow \infty$.
We first establish some well-known facts about random walks on graphs. First, $\mbf{P}^\infty$ is well-defined under our assumption that the graph is undirected, connected, and non-bipartite. It is rank-$1$ and equal to $\mbf{1} \bm{\tilde{d}}^\top$, where $\mbf{1}$ is a column vector of ones and $\bm{\tilde{d}}$ is the probability mass of each vertex in the stationary distribution of the random walk as a column vector. Note that $\bm{\tilde{d}}_i = \bv{D}_{ii}/v_G$.  That is, the probability mass of a vertex in the stationary distribution is proportional to its degree. We let $\mbf{\tilde{D}} = \bv{D}/v_G$ denote the diagonal matrix with entries  $\mbf{\tilde{D}}_{ii} = \bm{\tilde{d}}_i$.

Let $\lambda_i$ and $\mbf{w_i}$ be the $i^\text{th}$ eigenvalue and eigenvector of the symmetrized transition matrix $\mbf{\tilde{P}} = \bv{D}^{-1/2}\bv{A} \bv{D}^{-1/2}$. We have $\lambda_1=1$ and $\mbf{w_1}=\left( \sqrt{\bm{\tilde{d}}_1}, \dots, \sqrt{\bm{\tilde{d}}_n} \right)^\top$. From \cite{levinsoneigenvalue}, for any positive integer $k$,
\begin{align}\label{eq:diff} \mbf{P}^k = \mbf{P}^\infty + \sum_{j=2}^n \lambda_j^k \mbf{v_j} \mbf{v_j}^\top \mbf{\tilde{D}}, 
\end{align}
where $\mbf{v_i} = \mbf{\tilde{D}}^{-1/2} \mbf{w_i}$. We rewrite the expression in  \eqref{eq:diff} for $\mbf{P}^k$ and the expression \eqref{eq:netMF} of Qiu et al. for the PMI matrix, setting the negative sampling ratio $b$ to $1$ for the latter (i.e., one negative sample per positive sample):
\begin{align*}
\mbf{P}^k &= \mbf{1} \bm{\tilde{d}}^\top + \mbf{\tilde{D}}^{-1/2} \sum_{j=2}^n \lambda_j^k \mbf{w_j} \mbf{w_j}^\top \mbf{\tilde{D}}^{1/2} \quad \text{and} \\
\mbf{M_T} &= \log\left( T^{-1} \sum_{k=1}^T \mbf{P}^k \mbf{\tilde{D}}^{-1} \right) .
\end{align*}
Substituting the former into the latter, then rearranging the order of the summations and applying the geometric series formula yields
\begin{align*}
    \mbf{M_T} 
    &= \log\left( \mbf{1}\mbf{1}^\top + T^{-1} \mbf{\tilde{D}}^{-1/2} \left( \sum_{k=1}^T\sum_{j=2}^n \lambda_j^k \mbf{w_j} \mbf{w_j}^\top \right) \mbf{\tilde{D}}^{-1/2} \right) \\
    &= \log\left( \mbf{1}\mbf{1}^\top + T^{-1} \mbf{\tilde{D}}^{-1/2} \left( \sum_{j=2}^n \frac{\lambda_j (1 - \lambda_j^T)}{1 - \lambda_j} \mbf{w_j} \mbf{w_j}^\top \right) \mbf{\tilde{D}}^{-1/2}  \right) .
\end{align*}
Now we consider the limit as $T\to \infty$. Define
\begin{align*}
   \bv{M_\infty} = \lim_{T\to\infty}T \cdot \mbf{M_T}.
\end{align*}
Since $|\lambda_j| < 1$ for $j\neq 1$ \cite{levinsoneigenvalue}, the $(1 - \lambda_j^T)$ terms go to $1$ as $T \rightarrow \infty$ and we have:
\[ \bv{M_\infty} = \lim_{T\to\infty}T \cdot \log\left( \mbf{1}\mbf{1}^\top + T^{-1} \mbf{\tilde{D}}^{-1/2} \left( \sum_{j=2}^n \frac{\lambda_j}{1 - \lambda_j} \mbf{w_j} \mbf{w_j}^\top \right) \mbf{\tilde{D}}^{-1/2}  \right) .\] 
Since $\log(1+\epsilon) \to \epsilon$ as $\epsilon \to 0$, for any constant real number $c$, $T \log(1 + T^{-1}c) \to c$ as $T \to \infty$. We apply this identity element-wise, then simplify:
\begin{align*}
    \bv{M_\infty}
    &= \mbf{\tilde{D}}^{-1/2} \left( \sum_{j=2}^n \frac{\lambda_j}{1 - \lambda_j} \mbf{w_j} \mbf{w_j}^\top \right) \mbf{\tilde{D}}^{-1/2} \\
    &= \mbf{\tilde{D}}^{-1/2} \left( \sum_{j=2}^n \frac{1}{1 - \lambda_j} \mbf{w_j} \mbf{w_j}^\top - \sum_{j=2}^n \frac{1 - \lambda_j}{1 - \lambda_j} \mbf{w_j} \mbf{w_j}^\top \right) \mbf{\tilde{D}}^{-1/2} \\
    &= \mbf{\tilde{D}}^{-1/2} \left( \sum_{j=2}^n \frac{1}{1 - \lambda_j} \mbf{w_j} \mbf{w_j}^\top + \mbf{w_1} \mbf{w_1}^\top - \sum_{j=1}^n \mbf{w_j} \mbf{w_j}^\top \right) \mbf{\tilde{D}}^{-1/2} \\
    &= \mbf{\tilde{D}}^{-1/2} \left( \mbf{\tilde{L}}^+ + \mbf{w_1} \mbf{w_1}^\top - \mbf{I} \right) \mbf{\tilde{D}}^{-1/2} \\
    &= \mbf{1}\mbf{1}^\top + \mbf{\tilde{D}}^{-1/2} \left( \mbf{\tilde{L}}^+ - \mbf{I} \right) \mbf{\tilde{D}}^{-1/2} ,
\end{align*}
where the last step follows from $\mbf{w_1}$ being the element-wise square root of $\bm{\tilde{d}}$. Note that the above equation gives the expression for $\bv{M_\infty}$ in \eqref{eq:simple}, since $\bv{\tilde D} = \bv{D}/v_G$.
Similar analysis also leads to the following general expressions for the finite-$T$ PMI matrix:
%
\begin{align}\label{eq:finite}
    \hspace{-1em} \mbf{M_T}
    &= \log\left( \mbf{1}\mbf{1}^\top + T^{-1} \mbf{\tilde{D}}^{-1/2} \mbf{\tilde{P}} \mbf{\tilde{L}}^+ ( \mbf{I} - \mbf{\tilde{P}}^T ) \mbf{\tilde{D}}^{-1/2} \right) \nonumber \\
    = &\log\left( \mbf{1}\mbf{1}^\top + T^{-1} \left( \mbf{1}\mbf{1}^\top + \mbf{\tilde{D}}^{-1/2} \left( \mbf{\tilde{L}}^+ \left( \mbf{I} - \mbf{\tilde{P}}^{T+1} \right) - \mbf{I} \right) \mbf{\tilde{D}}^{-1/2} \right) \right)  .
\end{align}

\subsection{Approximation of Finite-$T$ PMI Matrix via Limiting PMI Matrix}\label{sec:infinite}
Note that the expression \eqref{eq:finite} above for the finite-$T$ matrix, when multiplied by $T^{-1}$ differs from the limiting matrix only by the term $\mbf{\tilde{D}}^{-1/2} \mbf{\tilde{L}}^+ \mbf{\tilde{P}}^{T+1} \mbf{\tilde{D}}^{-1/2}$, which vanishes as $T \to \infty$.
So, we may approximate the finite-$T$ matrix by simply dropping this term as follows:
\begin{align}\label{eq:approx}
    \mbf{M_T} 
    &\approx \log\left( R \left( \mbf{1}\mbf{1}^\top + T^{-1} \left( \mbf{1}\mbf{1}^\top + \mbf{\tilde{D}}^{-1/2} \left( \mbf{\tilde{L}}^+ - \mbf{I} \right) \mbf{\tilde{D}}^{-1/2} \right) \right) \right) \nonumber\\
    &= \log\left( R \left( \mbf{1}\mbf{1}^\top + T^{-1} \bv{M_\infty} \right) \right) ,
\end{align}
where $R(x)$ is any ramp function to ensure that the argument of the logarithm is positive. In our implementation, we use the function $R_\epsilon(x) = \max(\epsilon, x)$. We use the 64-bit floating-point machine precision limit ($\sim e^{-36}$) for $\epsilon$. Note that the NetMF method of \cite{qiu2018network} uses $R_1(x)$; we find that a small positive value consistently performs better. Both ramping functions can be interpreted as producing the positive shifted PMI matrix (shifted PPMI) matrix introduced by Levy and Goldberg~\cite{levy2014neural}. Other methods of avoiding invalid arguments to the logarithm are an interesting avenue for future work.

Note that the accuracy of \eqref{eq:approx}  is limited by the second largest eigenvalue of $\mbf{\tilde{P}}$, which is known as the Fiedler eigenvalue. Smaller magnitudes of this eigenvalue are correlated with a faster mixing rate~\cite{levinsoneigenvalue}, the rate at which $\mbf{P}^k \to \mbf{P}^\infty$ as $k$ increases. In Section \ref{sec:proc} we show that for typical graphs, the Fiedler eigenvalue is relatively small, and so the approximation is very accurate for large $T$, e.g., $T = 10$, which is a typical setting for DeepWalk. The approximation is less accurate for small $T$, e.g., $T = 1$ (See Table \ref{approx_error_table}.)

\paragraph{Effect of the Window Size $T$}

Intuitively, the effect of $T$ in \eqref{eq:approx} is to control the ``strength" of the logarithm nonlinearity, since, as noted previously, for any real constant $c$, $\log(1 + T^{-1}c) \to T^{-1}c$ as $T \to \infty$. That is, the logarithm becomes a simple linear scaling in this limit. As we will see, even when the approximation of \eqref{eq:approx} in inaccurate (when $T$ is very low) this approximated matrix qualitatively has similar properties to the actual $T$-window PMI matrix, and produces similar quality embeddings, as measured by performance in downstream classification tasks. This finding suggests that the strength of the logarithm nonlinearity can influence the locality of the embedding (as modulated in DeepWalk by the window size $T$) independently of modifying the arguments of the nonlinearity, which contain the actual information from the network as filtered by length-$T$ windows.

\subsection{Binarized Laplacian Pseudoinverse}\label{sec:binary}

Motivated by the view of DeepWalk as a variant of classic Laplacian factorization with an entrywise nonlinearity, we investigate a highly simplified version of InfiniteWalk. 
We construct a binary matrix by 1) computing the pseudoinverse of the unnormalized Laplacian $\bv{L}^+$, 2) picking a quantile hyperparameter $q \in (0,1)$, 3) determining the quantile $q$ element value, and 4) setting all elements less than this value to $0$ and others to $1$. In other words, an element of this matrix $\mbf{B}$ is given by $\mbf{B}_{ij} = [(\mbf{L}^+)_{ij} \geq c]$, where $c$ is the $q$ quantile element of $\mbf{L}^+$. We then produce embeddings by partially factoring this matrix $\mbf{B}$ as with the PMI matrices. Interestingly, this can be interpreted as factorizing the adjacency matrix of an implicit derived network whose sparsity is determined by $q$. Gaining a better understanding of the structure and interpretation of this network is an interesting direction from future work.

Note that in this method, we use the unnormalized Laplacian $\mbf{L}$ rather than the normalized Laplacian $\mbf{\tilde{L}} = \mbf{D}^{-1/2} \mbf{L} \mbf{D}^{-1/2}$, which appears in the expression \eqref{eq:simple} for $\bv{M_\infty}$. This is because, as we will show, 
%
the limiting PMI matrix is equal to the pseudoinverse of the unnormalized Laplacian up to a rank-3 term and a diagonal adjustment. We can rewrite our expression for the limiting matrix by expanding the normalized Laplacian in terms of the normalized Laplacian as follows:
\[ \mbf{M_\infty} = \mbf{1}\mbf{1}^T + v_G \left( \underbrace{\mbf{D}^{-1/2} \left(\mbf{D}^{-1/2} \mbf{L} \mbf{D}^{-1/2} \right)^+  \mbf{D}^{-1/2}} - \mbf{D}^{-1} \right) . \]
Consider the underbraced term above containing $\mbf{L}$. If this term had a true inverse rather than a pseudoinverse, the four factors involving the degree matrix would simply cancel. Instead, application of a variant of the Sherman-Morrison formula for pseudoinverses~\cite{meyer1973generalized} results in the following expression for this term:
\[ \mbf{D}^{-1/2} \left(\mbf{D}^{-1/2} \mbf{L} \mbf{D}^{-1/2} \right)^+  \mbf{D}^{-1/2} = (\mbf{I} - \mbf{1} \bm{\tilde{d}}^\top) \mbf{L}^+ (\mbf{I} - \bm{\tilde{d}} \mbf{1}^\top) .\]
This yields the following alternate expression for the limiting PMI matrix:
\begin{align}\label{eq:rank3} \mbf{M_\infty} = \mbf{1}\mbf{1}^T + v_G \left( (\mbf{I} - \mbf{1} \bm{\tilde{d}}^\top) \mbf{L}^+ (\mbf{I} - \bm{\tilde{d}} \mbf{1}^\top) - \mbf{D}^{-1} \right) .
\end{align}
In our context of binarizing $\mbf{L}^+$ by a quantile, note that addition by the all-ones matrix and multiplication by $v_G$ does not affect the ranks of the elements within the matrix, and the subtraction by the diagonal matrix $\mbf{D}^{-1}$ affects relatively few elements. Hence we might expect binarizing $\mbf{L}^+$ by thresholding on quantiles to have a similar effect as binarizing the limiting PMI matrix itself.

Binarization is arguably one of the simplest possible methods of augmenting the Laplacian with a nonlinearity. As we will see, this method has good downstream performance compared to DeepWalk and related methods. We argue that this suggests that the core advancement of deep vertex embeddings over classical spectral embedding methods based on factorizing the Laplacian is application of a nonlinearity.

\section{Experimental Setup}

We now discuss how we empirically validate the performance of the limiting PMI matrix method presented in Section \ref{sec:infinite} and the Laplacian pseudoinverse binarization method of Section \ref{sec:binary}.

\subsection{Data Sets}

\begin{figure} 
\begin{center}
\includegraphics[width=1.0\linewidth]{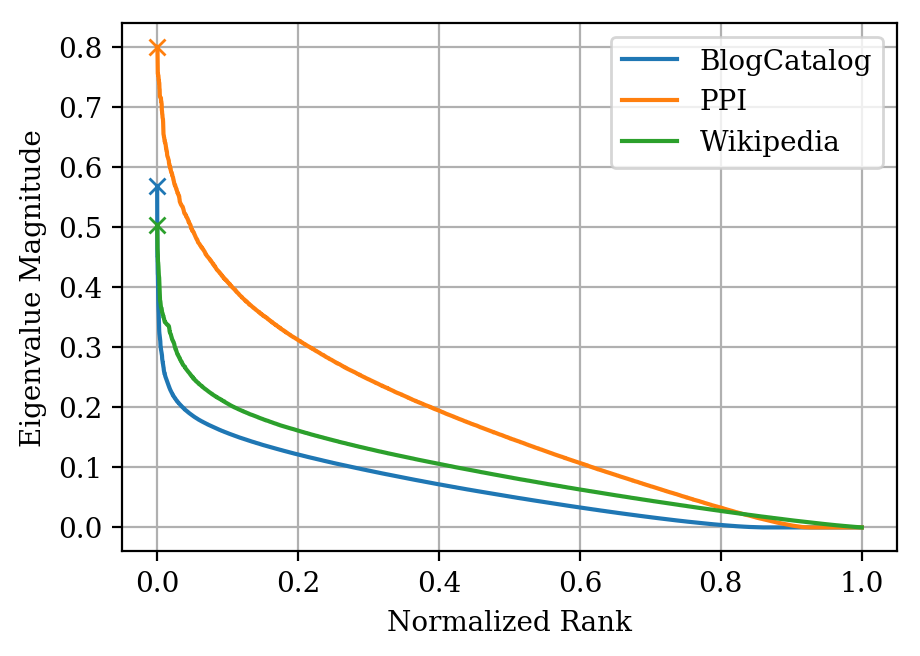}
\end{center}
\caption{Sorted eigenvalues of $\mbf{\tilde{P}}$ for each network.
The top eigenvalue of 1 is excluded, and the Fiedler eigenvalues are marked with X's.}
\label{prob_eigvals}
\end{figure}

We use three of the four datasets used in the evaluation of the NetMF algorithm \cite{qiu2018network}. Table~\ref{dataset_info_table} provides network statistics. Figure \ref{prob_eigvals} provides the eigenvalue distribution of the symmetrized random walk matrix $\mbf{\tilde{P}}$ for each network.

\textbf{\textsc{BlogCatalog}}~\footnote{\url{http://leitang.net/code/social-dimension/data/blogcatalog.mat}}~\cite{agarwal2009social} is a social network of bloggers. The edges represent friendships between bloggers, and vertex labels represent group memberships corresponding to interests of bloggers.

\textbf{\textsc{Protein-Protein Interaction (PPI)}}~\footnote{\url{http://snap.stanford.edu/node2vec/Homo_sapiens.mat}}~\cite{stark2010biogrid} is a subgraph of the \textsc{PPI} network for Homo Sapiens. Vertices represent proteins, edges represent interactions between proteins, and vertex labels represent biological states. We use only the largest connected component, which has over $99\%$ of the nodes.

\textbf{\textsc{Wikipedia}}~\footnote{\url{http://snap.stanford.edu/node2vec/POS.mat}} is a co-occurrence network of words from a portion of the Wikipedia dump. Nodes represent words, edges represent co-occurrences within windows of length $2$ in the corpus, and labels represent inferred part of speech (POS) tags.


\begin{table}[t]
\begin{center}
\begin{tabular}{c|c|c|c}
\hline
\hline
Dataset   & \textsc{BlogCatalog} & \textsc{PPI}  & \textsc{Wikipedia} \\ \hline
$|V|$       & 10,312      & 3,852            & 4,777     \\ \hline
$|E|$       & 333,983     & 76,546           & 184,812   \\ \hline
Fiedler Eigenvalue & 0.568          & 0.800                  & 0.504        \\ \hline
\# Labels & 39          & 50                  & 40        \\ \hline
\hline
\end{tabular}
\smallskip
\caption{Network Statistics.}
\label{dataset_info_table}
\end{center}
\end{table}

\subsection{Procedure}\label{sec:proc}

\paragraph{Implementation} See Algorithm \ref{alg:alg} for the pseudocode of our limiting PMI matrix method. We use the NumPy~\cite{oliphant2006guide} and SciPy~\cite{scipy} libraries for our implementation. The most expensive component of the algorithm is the pseudoinversion of the graph Laplacian. While there is significant literature on approximating this matrix, or vector products with it \cite{spielman2004nearly,koutis2014approaching,kelner2013simple,ranjan2014incremental}, we simply use the dense SVD-based function from NumPy. For graphs of larger scale, this method would not be practical. The truncated sparse eigendecomposition is handled via SciPy's packer to ARPACK~\cite{lehoucq1998arpack}, which uses the Implicitly Restarted Lanczos Method. As in \cite{qiu2018network}, to generate a $d$-dimensional embedding, we return the singular vectors corresponding to the $d$ largest singular values, scaling the dimensions of the singular vectors by the square roots of the singular values. For classification, we use the scikit-learn~\cite{pedregosa2011scikit} packer to LIBLINEAR~\cite{fan2008liblinear}.
Demo code for InfiniteWalk is available at \href{https://www.github.com/schariya/infwalk}{github.com/schariya/infwalk}.

\begin{algorithm}
  \caption{InfiniteWalk}
  \begin{algorithmic}[1]
    \State Compute $\mbf{{M}_\infty} = \mbf{1}\mbf{1}^\top + \mbf{\tilde{D}}^{-1/2} \left( \mbf{\tilde{L}}^+ - \mbf{I} \right) \mbf{\tilde{D}}^{-1/2}$
    \State Compute $\mbf{M} = \log\left( R_\epsilon \left( \mbf{1}\mbf{1}^\top + T^{-1} \mbf{M_\infty} \right) \right)$
    \State Rank-$d$ approximation by truncated eigendecomposition: $\mbf{M} \approx \mbf{V} \times \text{diag}(\mbf{w}) \times \mbf{V}^\top$
    \State \Return $\mbf{V} \times \text{diag}(\sqrt{|\mbf{w}|})$ as vertex embeddings
  \end{algorithmic}\label{alg:alg}
\end{algorithm}

\paragraph{Evaluation Setting} To investigate the usefulness and meaningfulness of the limiting PMI matrix, we evaluate the quality of embeddings generated by its truncated SVD after applying the element-wise ramp-logarithm described in Section \ref{sec:infinite}. For this task, we closely follow the same procedure as in \cite{perozzi2014deepwalk} and \cite{qiu2018network}. We use one-vs-rest logistic regression on the embeddings for multi-label prediction on the datasets. The classifiers employ L2  regularization with inverse regularization strength $C=1$. Classifiers are trained on a portion of the data, with the training ratio being varied from $10\%$ to $90\%$; the remainder is used for testing. As in \cite{perozzi2014deepwalk} and \cite{qiu2018network}, we assume that the number of labels for each test example is given. In particular, given that a vertex is assigned $k$ labels, the classifier predicts exactly the $k$ classes to which it assigned the highest probability. We use the mean scores over $10$ random splits of the training and test data for each training ratio. We evaluate the micro-F1 and macro-F1 scores of classifiers using our embedding.

\paragraph{Hyperparameter Choices} We compare our results to those of DeepWalk \cite{perozzi2014deepwalk}, LINE \cite{tang2015line}, and NetMF \cite{qiu2018network} as reported in \cite{qiu2018network}. The hyperparameters used for DeepWalk are the preferred default settings of its authors: window size $T=10$, walk length $L=40$, and number of walks starting from each vertex $\gamma=80$. Results from the second-order variant of LINE are reported. As the authors of NetMF report results for window sizes $T=1$ and $T=10$, we similarly report results for InfiniteWalk with $T=1$ and $T=10$. We expect the results of InfiniteWalk, as an approximation of the NetMF method in the high $T$ limit, to at least be roughly similar for the higher $T=10$ setting. We also include results with our limiting $T\rightarrow\infty$ matrix, though only for illustrative purposes. As the limiting matrix is essentially a simple linear transformation of the graph Laplacian's pseudoinverse, we expect embeddings derived thereof to perform relatively poorly. The entrywise nonlinearity seems to be critical. The embedding dimensionality is $128$ for all methods as in both \cite{perozzi2014deepwalk} and \cite{qiu2018network}. 
%
%

\subsection{Binarized Laplacian Pseudoinverse}
We implement and evaluate the simplified method of factorizing a binarized version of the unnormalized Laplacian pseudoinverse (described in Section \ref{sec:binary}) in the same way. We present results for the best values of quantile hyperparameter $q$ found by rough grid search. As with the window size $T$, the best value is expected to vary across networks. We compare to the performance of NetMF, the sampling methods LINE and DeepWalk, and classical methods - since the normalized and unnormalized Laplacians themselves both perform poorly on these tasks, we compare to factorizing the adjacency matrix itself. Again, since inverting and binarizing is one of the simplest possible nonlinearities to apply the Laplacian, good downstream performance suggests that the addition of a nonlinearity is the key advancement of deep node embeddings from classical embeddings of the Laplacian itself.

\section{Results}

We now discuss our experimental results  on both the limiting PMI-based algorithm and the simple Laplacian binarization algorithm.

\begin{table}
\begin{center}
\begin{tabular}{c|c|c|c}
\hline
\hline
Dataset   & \textsc{BlogCatalog} & \textsc{PPI}  & \textsc{Wikipedia} \\ \hline
Error ($T=1$)       & 2.456     & 2.588           & 1.355   \\ \hline
Error ($T=10$)       & 0.001273     & 0.04152           & 0.004892   \\ \hline
Ramped Elts ($T=1$) & 0.1834          & 0.1440                  & 0.08892        \\ \hline
Ramped Elts ($T=10$) & 0.0004901          & 0.002521                  & 0.0005943        \\ \hline
\hline
\end{tabular}
\end{center}
\smallskip
\caption{PMI Approximation Error. The first two rows give the Frobenius norm of the difference between the true PPMI matrix $\bv{M}_T$ and our approximation based on $\bv{M}_\infty$ (see \eqref{eq:approx}), divided by the norm of $\bv{M}_T$. The log-ramp nonlinearity with $R_1$, as used in the NetMF method, is applied to both matrices. The last two rows give the number of elements that are affected by the ramping component of the nonlinearity in one matrix but not the other, normalized by the size of the matrices.}
\label{approx_error_table}
\vspace{-2em}
\end{table}

\subsection{PMI Approximation Error} In Table \ref{approx_error_table}, we show how closely the PMI approximation given by \eqref{eq:approx} matches the true PMI matrix. We can see from Table \ref{dataset_info_table} that the Fiedler eigenvalues of our graphs are bounded away from $1$. Thus, as expected, the approximation of the finite-$T$ PMI matrix via the limiting matrix is quite close at $T=10$, but not so at $T=1$.
Additionally, at $T=10$, the elements which are affected by the ramping component of the nonlinearity are similar between our approximation and the true PMI matrix.
The accurate approximation at $T=10$ explains why InfiniteWalk performs similarly on downstream classification tasks. Interestingly, at $T=1$, InfiniteWalk performs competitively, in spite of inaccurate approximation.

\subsection{Multi-Label Classification}
\begin{figure*}[p]
\begin{center}
\includegraphics[width=\linewidth]{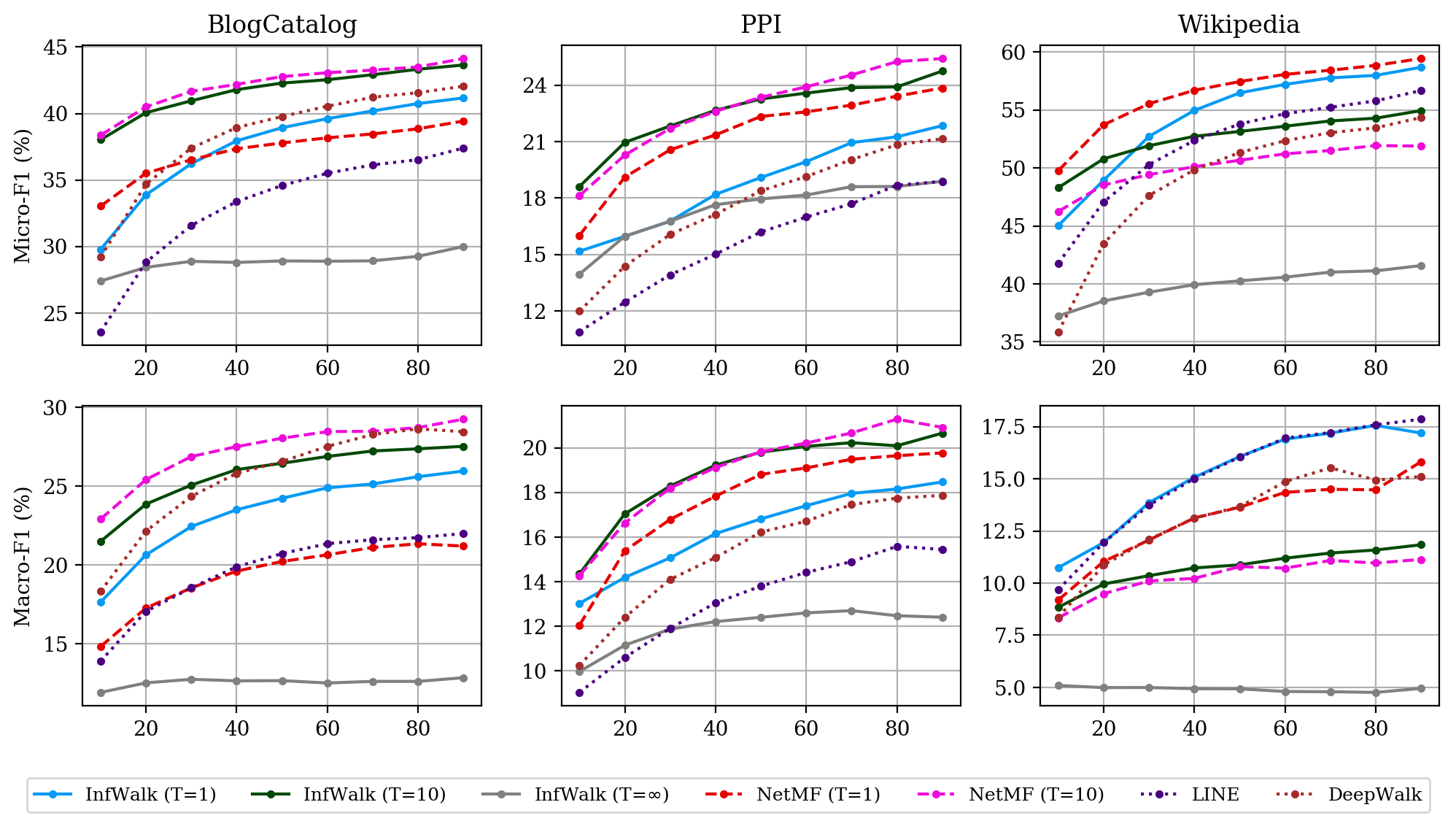}
\end{center}
\caption{Multi-label classification performance on the \textsc{BlogCatalog}, \textsc{PPI}, and \textsc{Wikipedia} networks. Micro-F1 score (top) and Macro-F1 score (bottom) versus percent of data used for training. Results for InfiniteWalk (Algorithm \ref{alg:alg}) all appear as solid lines.}
\label{all_f1_scores}
\end{figure*}

In Figure \ref{all_f1_scores} we show downstream performance of embeddings based on the limiting $\bv{M}_\infty$ approximation, compared to other methods. Across both metrics and all datasets, NetMF and InfiniteWalk are generally or par with or outperform the sampling-based methods, LINE and DeepWalk. As observed in \cite{qiu2018network}, DeepWalk and NetMF with $T=10$ have better overall performance than LINE and NetMF with $T=1$ on the \textsc{BlogCatalog} and \textsc{PPI} networks, while the inverse is true for the \textsc{Wikipedia} network. This suggests that shorter-range dependencies better capture \textsc{Wikipedia}'s network structure. As expected, InfiniteWalk with the $T=10$ nonlinearity performs better than the version with the $T=1$ nonlinearity on the former two datasets, while the inverse is true for \textsc{Wikipedia}. In all cases, the factorization of the $\bv{M_\infty}$ PMI matrix itself performs poorly. These findings support our hypothesis that changing the strength of the logarithm nonlinearity can largely emulate the effect of actually changing the window size $T$ in sampling and deterministic approaches.

\begin{figure*}[p]
\begin{center}
\includegraphics[width=0.95\linewidth]{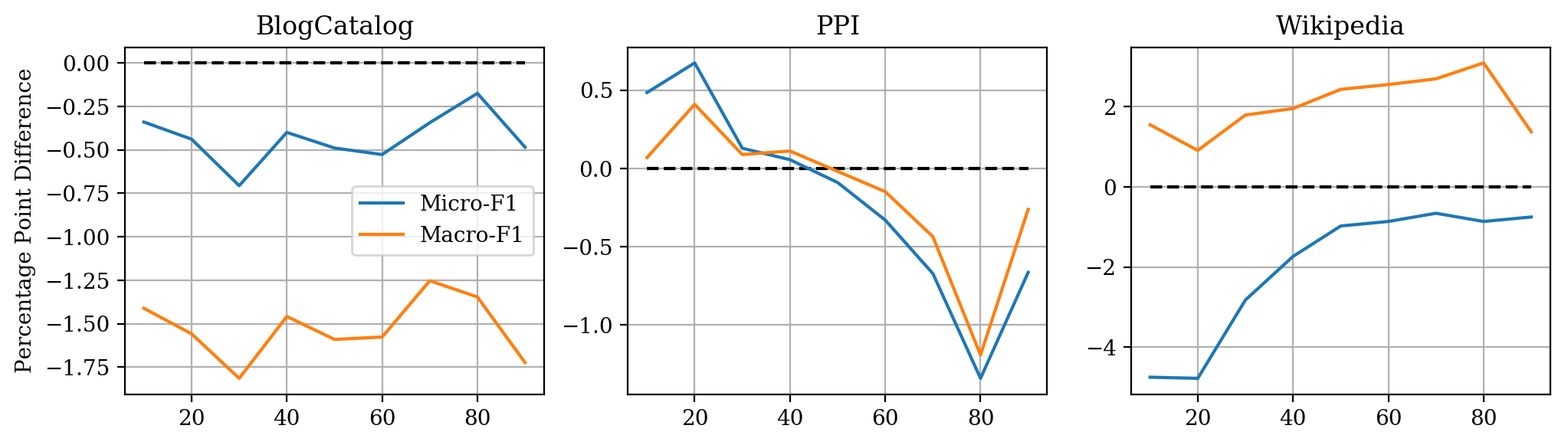}
\end{center}
\caption{Performance of InfiniteWalk relative to NetMF \cite{qiu2018network}. F1 score (\%) of InfiniteWalk minus F1 score of NetMF versus percent of data used for training. For both methods, $T=10$ is used for \textsc{BlogCatalog} and \textsc{PPI}, and $T=1$ is used for \textsc{Wikipedia}.}
\label{f1_compare}
\end{figure*}

While maximizing downstream performance is not the focus of our work, we observe that, overall, InfiniteWalk has performance competitive with if slightly inferior to NetMF (see Figure \ref{f1_compare}). On \textsc{BlogCatalog}, InfiniteWalk underperforms relative to NetMF. On \textsc{PPI}, InfiniteWalk outperforms NetMF when less than half the data is used for training, but underperforms when given more training data. On \textsc{Wikipedia}, InfiniteWalk underperforms relative to NetMF on macro-F1 score, but outperforms NetMF on micro-F1 score.

\begin{figure*}[p]
\begin{center}
\includegraphics[width=1.0\linewidth]{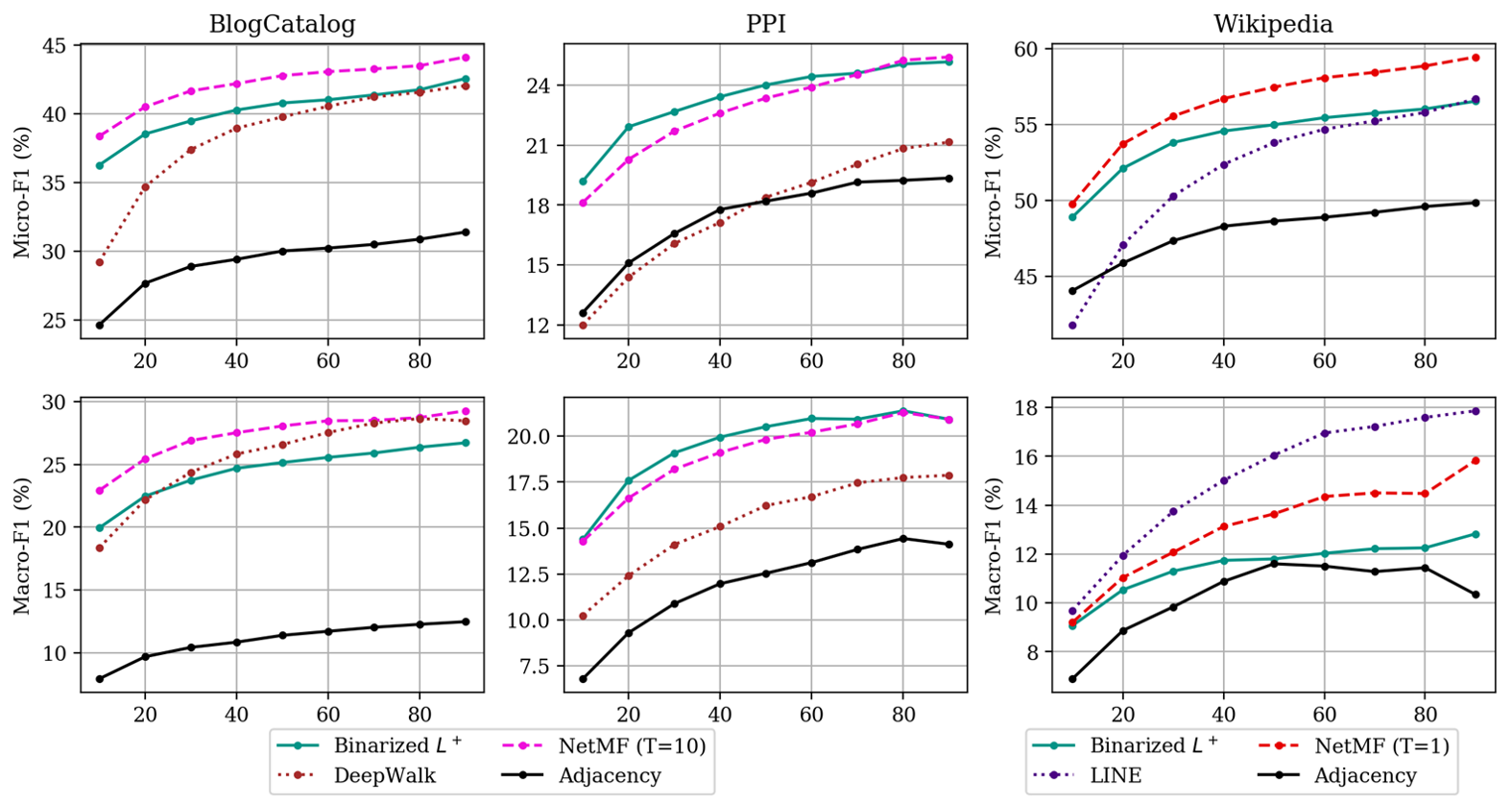}
\end{center}
\caption{Performance of the binarized Laplacian pseudoinverse method relative to NetMF, sampling-based methods, and embedding by factorizing the adjacency matrix. $0.95$ is used as the thresholding quantile for \textsc{BlogCatalog} and \textsc{PPI}, and $0.50$ is used for \textsc{Wikipedia}. The more suitable setting of $T$ and the more suitable sampling method is plotted for each network.}
\label{f1_bin}
\end{figure*}

\paragraph{Binarized Laplacian Pseudoinverse}
In Figure \ref{f1_bin} we show down stream performance of our simple method based on factorizing the binarized Laplacian pseudoinverse. This method performs remarkably well on both $T=10$ networks. On \textsc{PPI}, it matches the performance of NetMF, and on \textsc{BlogCatalog}, it is nearly on par again, accounting for most of the improvement from the classical method. On the $T=1$ network, \textsc{Wikipedia},
it is less successful, 
especially on Macro-F1 error, but still improves on the classical method. These result again support our hypothesis that the key ingredient of improved node embeddings seems to be the application of a nonlinearity to the Laplacian pseudoinverse.

\begin{figure*}[p]
\begin{center}
\includegraphics[width=1.0\linewidth]{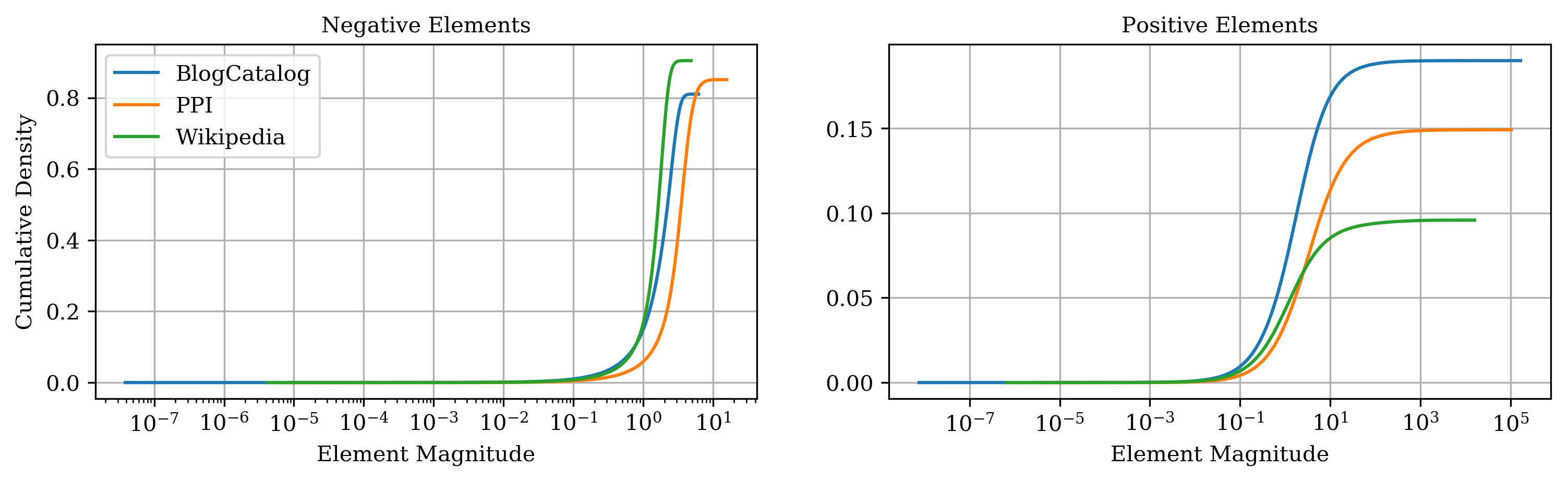}
\end{center}
\caption{Distribution of elements of the limiting PMI matrices $\bv{M}_\infty$ of the three networks. The distributions are separated between negative and positive elements corresponding to negative and positive correlations between nodes.}
\label{pmi_elements}
\end{figure*}

\paragraph{Elements of Limiting PMI Matrices}
Since we are introducing the limiting PMI matrix as an object for future investigation, we also give a preliminary qualitative description of its elements. See Figure \ref{pmi_elements} for a visualization of the element distribution for the three networks we investigate. Across these networks, these matrices tend to contain mostly small negative elements, corresponding to weak negative correlations between nodes, as well as some large positive values, corresponding to strong positive correlations. The distributions overall have similar shapes, and, interestingly, have roughly the same ratios of negative values to positive values, corresponding to roughly the same ratios of negative correlations to positive correlations.


\section{Discussion and Conclusions}

In this work we have simplified known expressions for the finite-$T$ network PMI matrix and derived an expression for the $T \to \infty$ matrix in terms of the pseudoinverse of the graph Laplacian. This expression strengthens connections between SGNS-based and class spectral embedding methods.

We show that, with a simple scaled logarithm nonlinearity, this limiting matrix can be used to approximate finite-$T$ matrices which yield competitive results on downstream vertex classification tasks. This finding  suggests that the core mechanism of SGNS methods as applied to networks is a simple nonlinear transformation of classical Laplacian embedding methods. We even find that just binarizing the Laplacian pseudoinverse by thresholding often accounts for most of the performance gain from classical approaches, suggesting again the important of the nonlinearity.

\paragraph{Future Work} 
We view our work as a step in understanding the core mechanisms of SGNS-based embedding methods. However many open questions remain. 

For example, one may ask why the scaled logarithm non-linearity is a good choice. Relatedly, how robust is performance to changes in the nonlinearity? Our results on binarization of the Laplacian pseudoinverse indicate that it may be quite robust, but this is worth further exploration. Finally, as discussed, our binarization method can be viewed as producing the adjacency matrix of a graph based on the Laplacian pseudoinverse, and then directly factoring this matrix. Understanding how this derived graph relates to the input graph would be a very interesting next step in understanding the surprisingly competitive performance of this method.

Additionally, as discussed previously, node2vec~\cite{grover2016node2vec} is a  generalization of DeepWalk that adds additional hyperparameters to create second-order random walks. Qiu et al.~\cite{qiu2018network} also provide an expression for the matrix that is implicitly factored by node2vec, so pursuing the $T\rightarrow\infty$ limit of this matrix may provide insight into node2vec and an interesting generalization of DeepWalk's limiting PMI matrix.

\bibliographystyle{ACM-Reference-Format}
\bibliography{sample-base}

\end{document}